\documentclass{article}
\usepackage{spconf,amsmath,graphicx}
\usepackage{cleveref}
\usepackage{tabularx}

\usepackage{graphicx}
\usepackage{amsmath}
\usepackage{amssymb}
\usepackage{booktabs}
\usepackage{wrapfig}
\usepackage{arydshln} 
\usepackage{makecell}
\usepackage{multirow}
\usepackage{pifont}

\usepackage{dsfont}
\usepackage{arydshln}
\usepackage{enumitem}
\title{Persona Extraction through Semantic Similarity for \\Emotional Support Conversation Generation}
\name{Seunghee Han, Se Jin Park, Chae Won Kim, Yong Man Ro\sthanks{Corresponding Author. This work was partially supported by the National Research Foundation of Korea (NRF) grant funded by the Korea government (MSIT) (No. NRF-2022R1A2C2005529) and BK21 FOUR (Connected AI Education \& Research Program for Industry and Society Innovation, KAIST EE, No. 4120200113769).}}
\address{Integrated Vision Language Lab, School of EE, KAIST, South Korea}
\begin{document}
\ninept
\maketitle
\begin{abstract}
Providing emotional support through dialogue systems is becoming increasingly important in today’s world, as it can support both mental health and social interactions in many conversation scenarios. Previous works have shown that using persona is effective for generating empathetic and supportive responses. They have often relied on pre-provided persona rather than inferring them during conversations. However, it is not always possible to obtain a user persona before the conversation begins. To address this challenge, we propose \textbf{PESS} (\textbf{P}ersona \textbf{E}xtraction through \textbf{S}emantic \textbf{S}imilarity), a novel framework that can automatically infer informative and consistent persona from dialogues. We devise completeness loss and consistency loss based on semantic similarity scores. The completeness loss encourages the model to generate missing persona information, and the consistency loss guides the model to distinguish between consistent and inconsistent persona. Our experimental results demonstrate that high-quality persona information inferred by PESS is effective in generating emotionally supportive responses.
\end{abstract}

\vspace{-0.1cm}

\section{INTRODUCTION}
\label{sec:intro}
\vspace{-0.1cm}
In today's world, many people feel isolated and have mental health difficulties due to the recent pandemic \cite{kumar2021covid, sikali2020dangers}. They may not have enough friends to turn to for help, or they may be afraid to seek help publicly. This circumstance  highlights the importance of dialogue systems and chatbots that can provide emotional support \cite{denecke2020mental, kraus2021towards}, providing a confidential space where people can express their feelings and seek help. Dialogue systems must get to know their users well to give appropriate emotional support. Getting to know users involves understanding their preferences, personality, and current state. This information is called the user's persona. Numerous studies \cite{liu-etal-2020-impress, cheng2022pal} have demonstrated the effectiveness of incorporating user persona into dialogue generation. A common approach in these studies has been to rely on pre-provided user persona, rather than inferring the persona during the conversation. However, in real-world settings, it is not always practical to ask users to provide their persona before the conversation begins.

To address this challenge, there is a need for models that can automatically infer persona information from dialogues. Several attempts have been made to extract persona information from dialogues. Xu \textit{et al.} \cite{xu2022long} trained classifiers to filter utterances that contain persona information from dialogue history. However, this approach is not always reliable, as user utterances can contain noise that makes it difficult to capture crucial persona information. Additionally, the user's persona may be expressed across multiple utterances, requiring a certain level of reasoning.
Another approach \cite{cheng2022pal} employed BART \cite{lewis-etal-2020-bart} which is fine-tuned on the Persona-Chat dataset \cite{zhang2018personalizing}. Given the user's entire utterances in dialogue history, their model is trained to infer the user's persona which consists of multiple persona sentences. However, their model often missed some persona information or generated some persona that didn't match with ground-truth persona. This is due to the lack of focus on ensuring sentence-level consistency between the inferred persona and the ground-truth persona.
Therefore, we need a more fine-grained approach that can take into account the sentence-level consistency of the persona.

We propose \textbf{PESS} (\textbf{P}ersona \textbf{E}xtraction through \textbf{S}emantic \textbf{S}imilarity) that can infer informative and consistent persona from dialogues by providing fine-grained signals to the model based on semantic similarity scores. We compare the ground-truth persona and generated persona at the sentence level to check if any persona sentences are missing and to determine whether each generated persona sentence is consistent or inconsistent with the ground-truth. The consistency of the generated persona with the ground-truth is determined by measuring semantic similarity scores between them. Based on this information, we design two losses Completeness Loss and Consistency Loss. The completeness loss encourages the model to identify and fill in missing persona information by penalizing the model for missing persona sentences. The consistency loss guides the model to distinguish between consistent and inconsistent persona by pulling the consistent persona sentences and the ground-truth persona closer, and by pushing the inconsistent persona sentences further away from the ground-truth persona.

In summary, our contributions are as follows: (1) We propose \textbf{PESS}, a persona extraction framework that can generate an informative and consistent persona. We achieve this by introducing Completeness Loss and Consistency Loss based on semantic similarity scores. The completeness loss is designed to encourage the generation of missing persona information, and the consistency loss guides the model to differentiate between consistent and inconsistent persona. (2) Our experiments demonstrate that the proposed persona extractor PESS generates a high-quality and consistent persona, and show that the persona inferred by PESS significantly contributes to generating emotionally supportive responses.

\vspace{-0.2cm}

\section{METHOD}
\label{sec:method}
\vspace{-0.1cm}
We start from comparing ground-truth persona and generated persona by measuring semantic similarity scores between them (\ref{ssec:semantic sim}). Then we train persona extractor using two key components: Completeness Loss (\ref{ssec:completeness loss}) and Consistency Loss (\ref{ssec:consistency loss}) based on the information derived in \Cref{ssec:semantic sim}. Next, we infer the user's persona from the dialogue history using the trained persona extractor and the inferred persona is leveraged to generate a response (\ref{ssec:response genration}). 

\subsection{Problem Definition}
\label{ssec:definition}
The conversation between two speakers A and B are represented as $\mathcal{U} = \{u^A_1, u^B_1, u^A_2, u^B_2 \cdots, u^A_n\}$, where $u^A_i$ and $u^B_i$ are the $i$-th utterances of speakers A and B, respectively, and $n$ is the number of conversation turns. The ground-truth personas of speakers A and B are denoted as $\mathcal{P}_A = \{p_{A_1}, p_{A_2}, \cdots, p_{A_{m}}\}$ and $\mathcal{P}_B = \{p_{B_1}, p_{B_2}, \cdots, p_{B_{r}}\}$. Here, $p_{A_i}$ and $p_{B_i}$ are $i$-th ground-truth persona sentences of each speaker, and $m$, $r$ are the number of speaker's persona sentences, respectively. Assume that the set of speaker A's utterances in $\mathcal{U}$ is $\mathcal{U}_A = \{u^A_1, u^A_2, \cdots, u^A_n\}$. Then the persona of speaker A, inferred from $\mathcal{U}_A$ by the persona extractor PESS is expressed as $\mathcal{P}^g_A=\text{PESS}(\mathcal{U}_A)=\{p^g_{A_1}, p^g_{A_2}, \cdots, p^g_{A_k}\}$, where $k$ is the number of the generated persona sentences. Here, the persona extractor is a transformer-based language model. We aim to encourage the persona extractor to generate an informative and consistent persona $\mathcal{P}^g_A$ from the speaker's utterances history $\mathcal{U}_A$. Then, we use the inferred persona $\mathcal{P}^g_A$ and dialogue history $\mathcal{U}$ to generate an appropriate response $u^B_n$, which is speaker B's response to speaker A's $n$-th utterance $u^A_n$. For the sake of brevity, in the rest of this paper, we refer to $\mathcal{P}_A$ as $\mathcal{P}= \{p_1, p_2, \cdots, p_{m}\}$ and $\mathcal{P}^g_A$ as $\mathcal{P}^g= \{p^g_1, p^g_2, \cdots, p^g_{k}\}$.

\begin{figure}[t!]
\centering
\includegraphics[width=0.9\linewidth]{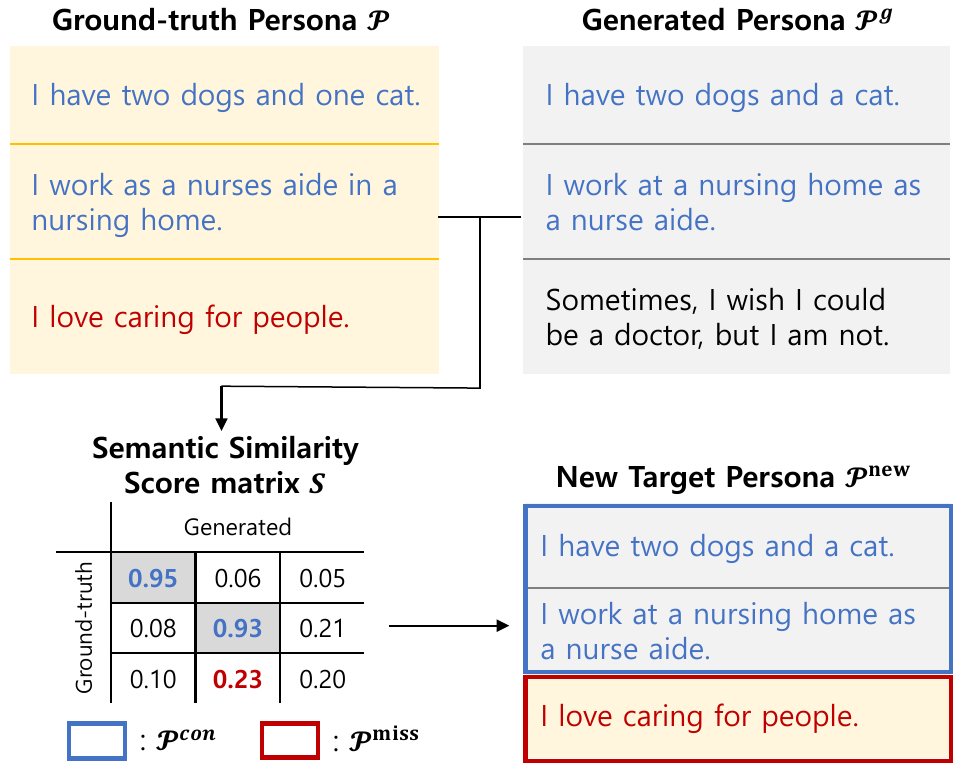}
\caption{The overall process of obtaining the consistent persona $\mathcal{P}^\text{con}$ and the missing persona $\mathcal{P}^\text{miss}$ based on semantic similarity score matrix $\mathcal{S
}$. New target persona $\mathcal{P}^\text{new}$ is constructed by combining $\mathcal{P}^\text{con}$ and $\mathcal{P}^\text{miss}$.} 
\label{fig:new target}
\end{figure}

\subsection{Semantic Similarity for Consistency Measurement}
\label{ssec:semantic sim}
\Cref{fig:new target} illustrates the overall process of finding the consistent persona and the missing persona. First, we compare the ground-truth persona and the generated persona. The generated persona may contain both sentences that are consistent and inconsistent with the ground-truth. Therefore, fine-grained comparison is needed to determine which sentences are consistent and which are not. Huang \textit{et al.} \cite{huang-etal-2023-swing} segment the generated summaries into sentences to determine faithfulness of each sentence. Inspired by the work of \cite{huang-etal-2023-swing}, we split the generated persona and the ground-truth persona into sentences and measure semantic similarity between each sentence pair. This gives us a semantic similarity score matrix $S \in \mathbb{R}^{m \times k}$. The semantic similarity score $s_{i,j}$ of $i$-th ground-truth persona sentence $p_i$ and $j$-th generated persona sentence $p^g_j$ is:
\begin{equation}
s_{i,j} = \text{sim}(E(p_i),E(p^g_j)),
\end{equation}
where $\text{sim}(\cdot,\cdot)$ is cosine similarity function, and $E$ represents sentence transformer \cite{ni-etal-2022-sentence} that calculates embedding of sentence.

Next, for each ground-truth persona sentence $p_i$, we find the best matching generated persona sentence $p^g_{j}$, which has maximum similarity score $s_{i,j}$. We denote the mapping function from the $i$-th ground-truth persona sentence to the best matching generated persona sentence as:
\begin{equation}
    I(i) = {\text{argmax}}_j s_{i,j}\,.
\end{equation}
Here, if $s_{i,I(i)}\geq \tau$ for a given ground-truth persona sentence $p_i$, it means that $p^g_{I(i)}$ is consistent with $p_i$. Otherwise, $p^g_{I(i)}$ is inconsistent with $p_i$, and there is no matching generated sentence for $p_i$. Here, $\tau$ is a hyperparameter that indicates similarity threshold. 

We refer to the set of consistently generated persona sentences as $\mathcal{P}^\text{con} = \{p^g_{I(i)}\mid s_{i,I(i)}\geq \tau \}$, and the set of missing ground-truth persona sentences is denoted as $\mathcal{P}^\text{miss} = \{p_i\mid \forall j \, s_{i,j} < \tau \}$. We utilize these two sets $\mathcal{P}^\text{con}$ and $\mathcal{P}^\text{miss}$ to design the completeness loss and the consistency loss.

\subsection{Completeness Loss}
\label{ssec:completeness loss}
To encourage the persona extractor to generate missing persona information, we construct a new target persona $\mathcal{P}^\text{new}$ in place of original target $\mathcal{P}$. An example is shown in \Cref{fig:new target}. $\mathcal{P}^\text{new}$ is formed by taking the union of the consistently generated persona $\mathcal{P}^\text{con}$ and the missing persona $\mathcal{P}^\text{miss}$: $ \mathcal{P}^\text{new} = \mathcal{P}^\text{con} \cup \mathcal{P}^\text{miss}.$

The completeness loss is standard negative log-likelihood loss on new target persona $\mathcal{P}^\text{new}$ as:
\begin{equation}
    \mathcal{L}_\text{complete} = -\sum_{t}\log p(\mathcal{P}^\text{new}_t \mid \mathcal{P}^\text{new}_{<t},\mathcal{U}_A),
\end{equation}
By introducing the completeness loss, we can provide additional fine-grained signals to the model. Specifically, we give a small penalty for consistent persona $\mathcal{P}^\text{con}$ since the model already generates them with a high probability. In contrast, a larger penalty is applied to missing persona $\mathcal{P}^\text{miss}$ due to its lower likelihood of being generated. This encourages the model to focus more on generating the missing persona information.

\subsection{Consistency Loss}
\label{ssec:consistency loss}
The objective of consistency loss is to train the persona extractor to distinguish between consistent and inconsistent persona sentences through contrastive learning \cite{gunel2020supervised}. The consistency loss guides the model to generate persona that are consistent with the ground-truth persona by pulling the consistent persona sentences and the ground-truth persona closer, and by pushing the inconsistent persona sentences further away from the ground-truth persona. Positive samples are persona sentences from consistently generated persona $\mathcal{P}^\text{con}$, and negative samples are persona sentences that are in generated persona $\mathcal{P}^g$ but not in $\mathcal{P}^\text{con}$. The goal is to maximize the agreement of the positive samples and minimize the agreement of the negative samples in the embedding space. The consistency loss is expressed as follows:
\begin{equation}
    \mathcal{L}_\text{consist} = -\sum_{p^g_i\in\mathcal{P}^\text{con}}\log{\frac{\exp(\text{sim}(h_i,h_{\mathcal{P}}))}{\sum_{p^g_j\in\mathcal{P}^g}\exp(\text{sim}(h_j,h_{\mathcal{P}}))}} ,
\end{equation}
where $h_i$ and $h_j$ are the decoder’s last layer representations of generated persona $\mathcal{P}^g$, $h_{\mathcal{P}}$ is the decoder’s last layer representations of the ground-truth persona.

\begin{table*}[t!]
\vspace{-0.2cm}
\centering
\renewcommand{\arraystretch}{1.1}
\renewcommand{\tabcolsep}{6.0mm}
\resizebox{0.99\linewidth}{!}{
\begin{tabular}{lcccccccc}
\Xhline{3\arrayrulewidth}
\textbf{Model} & \textbf{B-1}$\uparrow$ & \textbf{B-2}$\uparrow$ & \textbf{B-3}$\uparrow$ & \textbf{B-4}$\uparrow$ & \textbf{R-1}$\uparrow$ & \textbf{R-2}$\uparrow$ & \textbf{R-L}$\uparrow$ & \textbf{BS}$\uparrow$ \\ 
\hline
PAL-PE            & 41.14        & 34.41        & 28.97         & 24.01        & 44.94        & 19.59        & 42.97        & 91.60       \\ 
\cdashline{1-9}
w/o $\mathcal{L}_\text{complete}$    & \textbf{43.09}        & \textbf{36.76}        & \textbf{31.59}        & \textbf{26.75}        & 45.24        & 21.96        & 44.12        & 91.75       \\
w/o $\mathcal{L}_\text{consist}$      & 41.27        & 34.78        & 29.5        & 24.70        & 45.24        & 20.75        & 43.46        & 91.74       \\ 
\textbf{PESS}           & 42.96        & 36.57        & 31.31        & 26.43        & \textbf{45.91}        & \textbf{22.05}        & \textbf{44.28}        & \textbf{91.84}       \\
\Xhline{3\arrayrulewidth}
\end{tabular}
}
\vspace{-0.1cm}
\caption{Automatic evaluation result of each persona extractor on Persona-Chat. PAL-PE refers to the persona extractor of PAL. w/o $\mathcal{L}_\text{complete}$ and w/o $\mathcal{L}_\text{consist}$ denote variants of PESS by ablating the corresponding loss. PESS outperforms PAL-PE and each loss component contributes to an overall improvement of metric scores.}
\label{table:auto extractor}
\vspace{-0.1cm}
\end{table*}
\begin{table*}[]
\centering
\renewcommand{\arraystretch}{1.1}
\renewcommand{\tabcolsep}{6.0mm}
\resizebox{0.9\linewidth}{!}{
\begin{tabular}{l|cll}
\Xhline{3\arrayrulewidth}
{\textbf{\#}}  & \multicolumn{3}{c}{\textbf{Utterances}}                                                                                                                     \\ \hline
1 & \multicolumn{3}{c}{One of the foods I like is tasty sushi.}                                                                                                 \\ \hdashline
2 & \multicolumn{3}{c}{I love hearing the rap hip hop music.}                                                                                                   \\ \hline
{\textbf{\#}}       & \multicolumn{1}{c|}{\textbf{Ground-truth persona}}   & \multicolumn{1}{c|}{\textbf{PESS}}  & \multicolumn{1}{c}{\textbf{PAL-PE}}   \\ \hline
1      & \multicolumn{1}{l|}{My favorite food is sushi.}      &\multicolumn{1}{l|}{ My favorite food is sushi.}  & \multicolumn{1}{l}{My favorite food is sushi.}                   \\ \hdashline
2           & \multicolumn{1}{l|}{{I listen to rap music.}} & \multicolumn{1}{l|}{My favorite music genre is hip hop.} &   \\ \hdashline
3  & \multicolumn{1}{c|}{}       & \multicolumn{1}{c|}{}    & \multicolumn{1}{c}{{Black hair is my favorite color.}}            \\ 
\Xhline{3\arrayrulewidth}
\end{tabular}
}
\vspace{-0.15cm}
\caption{Comparison between the inferred persona by PAL-PE and PESS. While PESS is able to infer the persona from the second utterance, PAL-PE misses it. The third persona generated by PAL-PE is inconsistent and does not correspond to any utterance.}
\label{table:persona missing}
\vspace{-0.2cm}
\end{table*}

\begin{figure}[t!]
\centering
\includegraphics[width=0.7\linewidth]{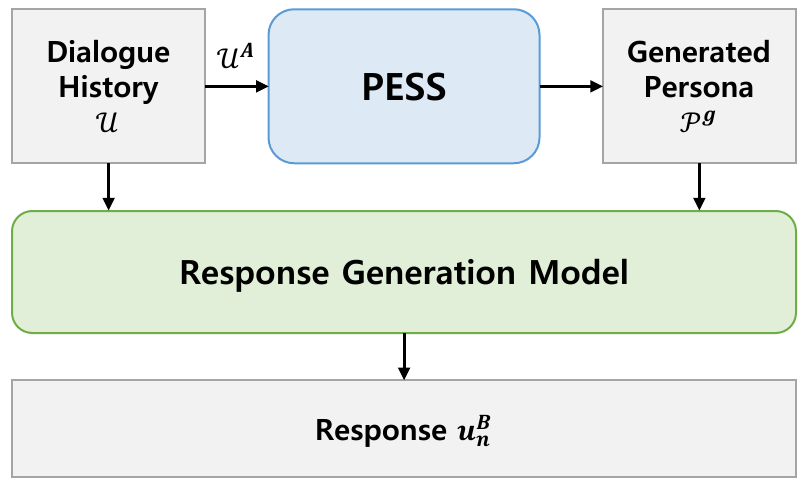}
\vspace{-0.2cm}
\caption{Overview of PESS-GEN. PESS generates the persona of speaker A, given the speaker A's utterances $\mathcal{U}_A$. The predicted persona $\mathcal{P}^g$ is passed to the response generation model with the dialogue history $\mathcal{U}$ to generate the speaker B's following response $u^B_n$.} 
\label{fig:response generation model }
\vspace{-0.2cm}
\end{figure}

\subsection{Training Objectives}
\label{training objectives}
The persona extractor \text{PESS} is trained with the combined total loss: 
\begin{equation}
    \mathcal{L} = \mathcal{L}_\text{nll} + \mathcal{L}_\text{complete} + \mathcal{L}_\text{consist},
\end{equation}
where $\mathcal{L}_\text{nll}$ is standard negative log-likelihood loss on the ground-truth persona $\mathcal{P}$, $\mathcal{L}_\text{nll} = -\sum_{t}\log p(\mathcal{P}_{t} \mid \mathcal{P}_{<t},\mathcal{U}_A).$

\subsection{Response Generation}
\label{ssec:response genration}
After training our persona extractor PESS, we freeze PESS and combine it with response generation model. We refer to this combined model as PESS-GEN. We adopt the architecture proposed in PAL \cite{cheng2022pal} for our response generation model. \Cref{fig:response generation model } illustrates the structure of PESS-GEN. Given dialogue history $\mathcal{U}$ which is a conversation between two speakers A and B, PESS infers speaker A's persona $\mathcal{P}^g$ from the set of speaker A's utterances $\mathcal{U}_A$.
Then, $\mathcal{P}^g$ and $\mathcal{U}$ are given to response generation model. The response generation model is trained to generate response $u^B_n$, which is speaker B's response to speaker A's last utterance. The training objective $\mathcal{L}_\text{gen}$ of response generation model is represented as:
\begin{equation}
    \mathcal{L}_\text{gen} = -\frac{1}{N}\sum^N_{t=1}\log p(u^B_{t} \mid u^B_{<t},\mathcal{U},\mathcal{P}^g),
\label{eq:7}
\end{equation}
where $N$ is length of $u^B_n$. We denote $u^B_n$ as $u^B$ in \Cref{eq:7}.

\begin{table*}[t!]
\vspace{-0.2cm}
\centering
\renewcommand{\arraystretch}{1.1}
\renewcommand{\tabcolsep}{5.0mm}
\resizebox{0.99 \linewidth}{!}{
\begin{tabular}{lccccccccc}
\Xhline{3\arrayrulewidth}
\textbf{Model}   & \textbf{ACC}$\uparrow$ & \textbf{PPL}$\downarrow$ & \textbf{B-1}$\uparrow$ & \textbf{B-2}$\uparrow$ & \textbf{B-3}$\uparrow$ & \textbf{B-4}$\uparrow$ & \textbf{D-1}$\uparrow$ & \textbf{D-2}$\uparrow$ & \textbf{R-L}$\uparrow$ \\ \hline
Blenderbot-joint & 27.72    & 18.11    & -        & 5.57     & -       & 1.93         & 3.74         & 20.66        & 16.36        \\
MISC             & 31.34    & 16.28    & -        & 6.60     & -       & 1.99         & 4.53         & 19.75        & 17.21        \\
GLHG             & -        & 15.67    & 19.66    & 7.57     & 3.74    & 2.13         & 3.50         & 21.61        & 16.37        \\ 
SUPPORTER        & -        & \textbf{15.37}    & 19.50    & 7.92     & 3.58    & -            & 4.93         & 27.73        & -        \\ 
PAL              & 32.58    & 17.66    & 19.92    & 7.88     & 4.08    & 2.45         & 5.06         & 28.94        & 16.68        \\ 
\hline
\textbf{PESS-GEN}   & \textbf{32.98}    & 16.92    & \textbf{21.03}    & \textbf{8.33}    & \textbf{4.28}     & \textbf{2.56}         & \textbf{5.28}         & \textbf{30.04}        & \textbf{17.49}        \\ 
\Xhline{3\arrayrulewidth}
\end{tabular}
}
\vspace{-0.1cm}
\caption{Automatic evaluation of response generation on ESConv dataset.}
\label{table:auto response}
\vspace{-0.1cm}
\end{table*}
\begin{table}[]
\centering
\renewcommand{\arraystretch}{1.1}
\resizebox{0.99 \linewidth}{!}{
\begin{tabular}{cl}
\Xhline{3\arrayrulewidth}
\multicolumn{2}{c}{\textbf{Dialogue history}}                                                                                                                                     \\ \hline
A   &  Well I'm distraught					
                                                                                              \\ \hdashline
B         &  Why do you feel that way?		  \\ \hdashline
A            &  Because my girlfriend has left me	   \\ \hdashline
\multicolumn{1}{c}{\multirow{2}{*}{B}} & That must be really hard for you. How long \\ 
\multicolumn{1}{c}{}                   &  were you together?\\ \hdashline
A            &  Four years					    \\ \hline
\multicolumn{2}{c}{\textbf{Response}}                                                                                                                                             \\ \hline
PAL               &  How did you two feel about it at that time?		        \\ \hdashline
\multicolumn{1}{c}{\multirow{3}{*}{\textbf{PESS-GEN}}} & How do you feel about that? \\ 
\multicolumn{1}{c}{}                   & Do you think that it would be a good thing to move \\
\multicolumn{1}{c}{}                   &  on from her? \\
\Xhline{3\arrayrulewidth}
\end{tabular}
}
\vspace{-0.1cm}
\caption{Comparison between dialogue response generated by PAL and PESS-GEN. Here, A is the seeker and B is the supporter.}
\label{tabel:res example}
\vspace{-0.1cm}
\end{table}
\begin{table}[t!]
\vspace{-0.1cm}
\centering
\renewcommand{\arraystretch}{1.1}
\renewcommand{\tabcolsep}{3.0mm}
\resizebox{0.99\linewidth}{!}{
\begin{tabular}{ccccc}
\Xhline{3\arrayrulewidth}
\textbf{Model}                & \textbf{Aspects} & \textbf{Win} & \textbf{Loss} & \textbf{Tie} \\ \hline
\multirow{5}{*}{PESS-GEN vs. PAL} & Flu.                & \textbf{40.6}         & 22.9          & 36.5         \\
                              & Ide.                & \textbf{43.8}         & 27.1          & 29.1         \\
                              & Com.                & \textbf{45.8}         & 31.3          & 22.9         \\
                              & Sug.                & \textbf{47.9}         & 24.0          & 28.1         \\
                              & Ove.                & \textbf{52.1}         & 34.4          & 13.5         \\ 
\Xhline{3\arrayrulewidth}
\end{tabular}}
\vspace{-0.1cm}
\caption{Human A/B evaluation on emotionally supportive responses generated by PESS-GEN and PAL (\%).}
\label{table:human}
\vspace{-0.4cm}
\end{table}

\section{EXPERIMENTS}
\label{sec:experiments}

\subsection{Setup}
\label{ssec:setup}
\textbf{Datasets.} We use Persona-Chat \cite{zhang2018personalizing} to train our persona extractor. Persona-Chat is a dataset that contains conversation of two participants and corresponding persona profiles. There are 8939 dialogues (65719 turns, 1015 personas) for training, 1000 dialogues (7801 turns, 100 personas) for validation, and 968 dialogues (7512 turns, 100 personas) for test. For response generation experiments, we use ESConv \cite{liu2021towards}. ESConv is a dataset of emotional support conversations between a help-seeker and a supporter. There is no persona information of them in this dataset. 1053 dialogues are split in a 7:2:1 ratio for training, validation, and test.\\
\textbf{Baselines.} In persona extractor evaluation, we compare our model with persona extractor of PAL which is BART \cite{lewis-etal-2020-bart} fine-tuned with a cross-entropy loss on Persona-Chat. For the sake of brevity, we refer to the persona extractor of PAL as PAL-PE in the experiment. In response generation evaluation, we compare our model with Blenderbot-Joint \cite{liu2021towards}, MISC \cite{tu2022misc}, GLHG \cite{peng2022control}, PAL \cite{cheng2022pal} and SUPPORTER \cite{zhou-etal-2023-facilitating}. Note that PAL has the same structure as our model PESS-GEN, except for the persona extractor. \\
\textbf{Implementation Details.} We chose a BART \cite{lewis-etal-2020-bart} pre-trained on CNN Daily Mail \cite{hermann2015teaching} as the backbone of our persona extractor PESS. Persona extractor is optimized for 8 epochs using AdamW with learning rate of 1e-5. For the first 4 epochs, persona extractor trained with only negative log-likelihood loss. Then it is trained with the proposed losses for the remaining epochs. We fix $\tau$ to 0.9 for all of our experiments. When we train response generation model, we follow setting of \cite{cheng2022pal}.\\
\textbf{Metrics.} We evaluate persona extractor PESS and response generation model PESS-GEN using following automatic evaluation: BLEU-n (B-n) \cite{10.3115/1073083.1073135}, Rouge-n (R-n), Rouge-L (R-L) \cite{lin-2004-rouge}, BERTScore (BS) \cite{zhang2019bertscore}, perplexity (PPL) \cite{bengio2000neural}, Distinct-n (D-n) \cite{li-etal-2016-diversity}. For Human A/B evaluation, 32 dialogues are randomly selected from the test set, and responses are generated using PESS-GEN and PAL. We ask 3 annotators to choose the better response (\textit{win}) for each dialogue. The annotators can also select a \textit{Tie} if the response from both models are equally good. Following \cite{liu2021towards}, we evaluate responses on five aspects: Fluency (Flu.), Identification (Ide.), Comforting (Com.), Suggestion (Sug.), and Overall (Ove.). 


\subsection{Persona Extractor Evaluation}
\label{ssec:persona extractor eval}
We compare PESS with PAL-PE, a persona extractor of PAL, on the Persona-Chat dataset. As shown in \Cref{table:auto extractor}, PESS outperforms PAL-PE on all automatic evaluation metrics, which means that the proposed method can infer more informative and consistent personas from dialogues than PAL-PE. To investigate the importance of completeness loss and consistency loss, we conducted an ablation study by removing them from the final training object. Both PESS w/o $\mathcal{L}_\text{complete}$ and PESS w/o $\mathcal{L}_\text{consit}$ show better performance compared to PAL-PE. This implies that the introduction of $\mathcal{L}_\text{complete}$ and $\mathcal{L}_\text{consit}$ improves persona extraction capabilities of model, respectively. When we introduce both losses, BLEU-n scores are slightly lower than w/o $\mathcal{L}_\text{complete}$. This is because BLEU-n scores are sensitive to changes in the words. Introducing $\mathcal{L}_\text{complete}$ may slightly decrease the BLEU-n scores because even if the words do not match exactly, if the semantic matches, it is considered a well-generated persona. However, since semantic similarity with ground-truth persona is much more important than exact matching of words, our PESS, which shows higher ROUGE-n,L score and BERTScore is more suitable for our purposes. \Cref{table:persona missing} shows an example of inferred persona by PAL-PE and PESS. While PESS successfully generates a persona ``My favorite music genre is hip hop." that matches the second ground-truth persona, PAL-PE misses this persona. This implies that our completeness loss encourages the persona extractor to generate missing persona. Moreover, while PAL-PE generates third persona doesn't correspond to any utterance, PESS doesn't generate inconsistent persona. This implies that the consistency loss successfully guides the model to distinguish between consistent and inconsistent persona.

\vspace{-0.1cm}
\subsection{Response Generation Evaluation}
\label{ssec:persona extractor}
\textbf{Automatic Evaluation}
We evaluate the response generation quality of baselines and PESS-GEN on ESConv dataset. In \Cref{table:auto response}, our PESS-GEN outperforms all baseline models on most evaluation metrics. The increase in BLEU and ROUGE scores indicates that our model generates responses similar to the human-produced reference sentences. Additionally, the high Distinct score suggests that our model generates diverse responses.
Considering the use of additional knowledge, response generation models can be divided into three groups: Blenderbot-Joint and SUPPORTER, which do not utilize additional knowledge; MISC and GLHG, which leverage common sense; and PAL and PESS-GEN, which use persona. Among these, the group that uses persona consistently performs better than the other groups. This is because persona provides more specific and relevant information about the seeker, which helps the model generate better responses.  Moreover, by comparing PAL and PESS-GEN we can demonstrate the importance of high-quality personas in generating emotionally supportive responses. 
\\
\textbf{Response Example.} We further compare the response generation results of PESS-GEN and PAL in Table \ref{tabel:res example}. PESS-GEN generates more empathetic and emotionally supportive responses than PAL. PESS-GEN recognizes the seeker's situation and attempts to empathize with the seeker. Also, PESS-GEN's response is more helpful, as it suggests the seeker to move forward.\\
\textbf{Human A/B Evaluation.}
In human evaluation, we compare our PESS-GEN with the current SOTA model PAL. As shown in \Cref{table:human}, PESS-GEN demonstrates superiority over PAL on all aspects. The difference is particularly noticeable in the suggestion aspect, indicating that the model is able to accurately identify the help-seeker's persona, including personality and needs, and provide appropriate suggestions. Furthermore, the large gain in the overall aspect indicates that high-quality persona inferred by PESS contributes significantly to generating an emotionally supportive response, not only in specific aspects but also overall.

\section{CONCLUSION}
\label{sec:conclusion}
In this paper, we introduce \textbf{PESS} (\textbf{P}ersona \textbf{E}xtraction through \textbf{S}emantic \textbf{S}imilarity), a framework that infers informative and consistent personas from dialogues. We propose completeness loss and consistency loss based on semantic similarity scores. The former loss encourages the model to identify and fill in missing persona information, while the latter loss guides the model to distinguish between consistent and inconsistent persona sentences. Experimental results demonstrate that our proposed persona extractor infer high-quality and consistent personas from dialogue history, which leads to generating emotionally supportive and appropriate response.

\vfill\pagebreak

\bibliographystyle{IEEEbib}
{\linespread{1.08}\selectfont\bibliography{refs}}

\end{document}